\documentclass[sigconf]{acmart}

\usepackage{graphicx}
\usepackage{amsmath}
\usepackage{amsthm}

\usepackage{amssymb}
\usepackage{booktabs} 
\usepackage{epstopdf} 
\usepackage{verbatim} 
\usepackage{subfigure} 
\usepackage{multirow} 
\usepackage{adjustbox} 
\usepackage{xcolor}
\usepackage{array}
\usepackage{enumitem}
\usepackage{bm}

\copyrightyear{2022}
\acmYear{2022}
\setcopyright{acmcopyright}\acmConference[KDD '22]{Proceedings of the 28th ACM SIGKDD Conference on Knowledge Discovery and Data Mining}{August 14--18, 2022}{Washington, DC, USA}
\acmBooktitle{Proceedings of the 28th ACM SIGKDD Conference on Knowledge Discovery and Data Mining (KDD '22), August 14--18, 2022, Washington, DC, USA}
\acmPrice{15.00}
\acmDOI{10.1145/3534678.3539340}
\acmISBN{978-1-4503-9385-0/22/08}

\begin{document}

\title{Label-enhanced Prototypical Network with Contrastive Learning for Multi-label Few-shot Aspect Category Detection}

\author{Han Liu}
\affiliation{
\institution{Dalian University of Technology}
\city{Dalian}
\country{China}}
\email{liu.han.dut@gmail.com}

\author{Feng Zhang}
\affiliation{
\institution{Peking University}
\city{Beijing}
\country{China}}
\email{zhangfeng@stu.pku.edu.cn}

\author{Xiaotong Zhang}
\authornote{Corresponding author.}
\affiliation{
\institution{Dalian University of Technology}
\city{Dalian}
\country{China}}
\email{zxt.dut@hotmail.com}

\author{Siyang Zhao}
\affiliation{
\institution{Dalian University of Technology}
\city{Dalian}
\country{China}}
\email{zhao\_siyang@mail.dlut.edu.cn}

\author{Junjie Sun}
\affiliation{
\institution{Dalian University of Technology}
\city{Dalian}
\country{China}}
\email{sunjunjiedlut@hotmail.com}

\author{Hong Yu}
\affiliation{
\institution{Dalian University of Technology}
\city{Dalian}
\country{China}}
\email{hongyu@dlut.edu.cn}

\author{Xianchao Zhang}
\authornotemark[1]
\affiliation{
\institution{Dalian University of Technology}
\city{Dalian}
\country{China}}
\email{xczhang@dlut.edu.cn}

\renewcommand{\shortauthors}{Han Liu et al.}

\begin{abstract}
Multi-label aspect category detection allows a given review sentence to contain multiple aspect categories, which is shown to be more practical in sentiment analysis and attracting increasing attention. As annotating large amounts of data is time-consuming and labor-intensive, data scarcity occurs frequently in real-world scenarios, which motivates multi-label few-shot aspect category detection. However, research on this problem is still in infancy and few methods are available. In this paper, we propose a novel label-enhanced prototypical network (LPN) for multi-label few-shot aspect category detection. The highlights of LPN can be summarized as follows. First, it leverages label description as auxiliary knowledge to learn more discriminative prototypes, which can retain aspect-relevant information while eliminating the harmful effect caused by irrelevant aspects. Second, it integrates with contrastive learning, which encourages that the sentences with the same aspect label are pulled together in embedding space while simultaneously pushing apart the sentences with different aspect labels. In addition, it introduces an adaptive multi-label inference module to predict the aspect count in the sentence, which is simple yet effective. Extensive experimental results on three datasets demonstrate that our proposed model LPN can consistently achieve state-of-the-art performance.
\end{abstract}



\maketitle

\section{Introduction}
Aspect Category Detection (ACD) is a fundamental task in sentiment analysis, which aims to identify the aspect categories mentioned in a given review sentence from a predefined aspect category set. As human usually make comments from different angles, i.e., a review sentence always contains multiple aspects, multi-label aspect category detection task came into existence. Existing approaches for multi-label ACD have achieved impressive and promising performance \cite{DBLP:conf/emnlp/LiYZP20, DBLP:conf/emnlp/HuZZCSCS19}. However, they rely heavily on large amounts of labeled data for each aspect. As annotating data is usually time-consuming, labor-intensive and even unachievable in real-world application, which motivates the multi-label few-shot aspect category detection task. 

Few-shot learning can recognize novel categories effectively with only a handful of labeled samples by exploiting the prior knowledge learned from previous categories, which is promising to break the data-shackles. Recent methods have made great progress in computer vision domain \cite{DBLP:conf/cvpr/WangXLZF20, DBLP:conf/iccv/0001HMKALK19} and natural language processing domain \cite{DBLP:conf/emnlp/HanZYWYLS18, DBLP:conf/emnlp/GaoHZLLSZ19}. Among these methods, prototypical network \cite{DBLP:conf/nips/SnellSZ17} is a powerful and potential model, which follows the episode learning strategy and uses the $N$-way $K$-shot setting. Specifically, in each episode, prototypical network first learns each prototype by averaging the corresponding $K$ support samples, and then predicts the labels of query samples based on the negative Euclidean distance between query samples and $N$ prototypes in support set.

\begin{figure}[t] 
	\centering
	\includegraphics[height=0.31\textheight,width=0.47\textwidth]{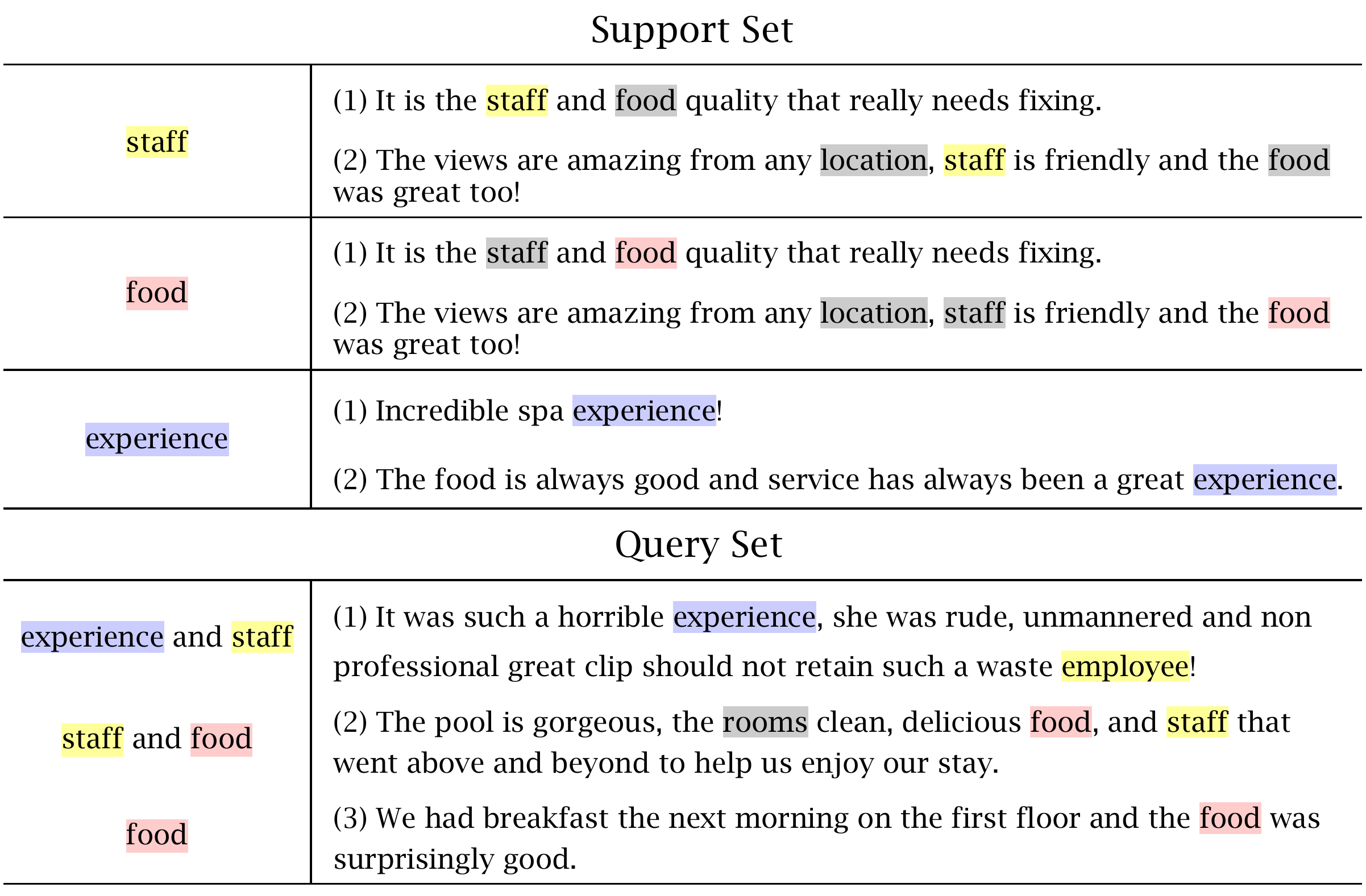}
	\caption{A meta-task example in 3-way 2-shot setting. The first column denotes the aspect label and the second column denotes the corresponding review sentence. As each review sentence may contain multiple aspects, we use different color background to mark the key words. The words in gray describe irrelevant (noisy) aspects, and the words in other colors represent the target aspects.} 
	\label{example} 
\end{figure}

Intuitively, we can directly extend prototypical network to solve the multi-label few-shot aspect category detection problem. However, there exist several challenging issues. (1) Simply calculating the prototype by averaging intra-class support samples may cause that different aspects share an identical prototype. Take an example in Figure \ref{example}, we construct an episode under 3-way 2-shot setting, i.e., there are 3 aspects with 2 samples per aspect in support set. As a review sentence may contain multiple aspects, it is quite possible that different aspects distribute in the same support samples. Here "staff" and "food" have the same support samples. Thus we will obtain the same prototype for "staff" and "food". Although the work \cite{DBLP:conf/acl/HuZGXGGCS20} attempts to use the attention mechanism to alleviate this issue, we find that it still does not work in this case. (2) When learning the prototype for a target aspect in multi-label scenarios, as each sentence probably contains several aspects, some irrelevant aspects will inevitably disturb the learning procedure. For example in Figure \ref{example}, when learning the prototype for "staff", the irrelevant aspects like "food" and "location" in  "The views are amazing from any location, staff is friendly and the food was great too!" will cause some negative impact. (3) Due to the diversity of human expression, different sentences may contain different numbers of aspects, so it is urgently needed to design an effective model to automatically predict the number of aspects in a sentence.  

In this paper, we propose a novel label-enhanced prototypical network (LPN) for multi-label few-shot aspect category detection. The main contributions of LPN consist of three parts. (1) By utilizing the label text description as complementary information to calculate the relationship between sentences and aspect labels and then obtain more discriminative prototypes, the LPN model can not only avoid the issue that different aspects share an identical prototype, but also retain aspect-relevant information while eliminating the negative effect triggered by irrelevant (noisy) aspects. (2) By integrating contrastive learning to obtain more powerful embeddings, the LPN model can push the embeddings from the same class close and embeddings from different classes further apart, thus facilitating the downstream aspect category detection task. (3) By introducing the adaptive multi-label inference module, the LPN model can determinate the number of aspects accurately. To verify the effectiveness of our proposed model, we conduct extensive experiments on three datasets. The empirical study shows that LPN can achieve state-of-the-art performance in comparison with other strong baselines.

\section{Related Work}
\subsection{Aspect Category Detection} 
Aspect Based Sentiment Analysis (ABSA) \cite{DBLP:journals/jis/ThetNK10} is a fine-grained sentiment analysis task that aims to extract aspects and predict the sentiment of each aspect. Aspect category detection (ACD) is an important subtask of ABSA, which aims to categorize a given review sentence into a set of predefined aspects. Previous studies mainly focus on single-aspect category detection, which include unsupervised and supervised methods. Unsupervised methods use semantic association analysis based on point-wise mutual information \cite{DBLP:conf/iccpol/SuXWSY06} or co-occurrence association rule mining \cite{DBLP:journals/tcyb/SchoutenWFD18} to extract aspects. These methods require a large amount of corpus resources and the performance is also barely satisfactory. Supervised methods exploit representation learning \cite{DBLP:conf/aaai/ZhouWX15}, topic-attention network \cite{DBLP:journals/corr/abs-1901-01183} or multilingual ngram-based convolutional network \cite{DBLP:conf/aaai/GhaderyMFS19} to identity different aspect categories. These methods have shown promising results in practice, but they rely heavily on a considerable amount of labeled data for each aspect to train a discriminative classifier. Due to the diversity and casualness of human expression, a review sentence often contains multiple aspects, which motivates multi-aspect category detection. Existing approaches for multi-label ACD \cite{DBLP:conf/emnlp/LiYZP20, DBLP:conf/emnlp/HuZZCSCS19} have achieved impressive performance. However, similar with supervised methods for single-aspect category detection, they also suffer from the serious data scarcity issue.

\subsection{Few-shot Learning}
Few-shot learning is a paradigm for solving the data deficiency problem, which aims to use the knowledge learned from seen classes, of which abundant labeled samples are available for training, to recognize unseen classes, of which limited labeled samples are provided. It has drawn much attention in computer vision domain \cite{DBLP:conf/cvpr/WangXLZF20, DBLP:conf/iccv/0001HMKALK19} and natural language processing domain \cite{DBLP:conf/naacl/YuHZDPL21, DBLP:conf/slt/0007KGLAG21}. Meta-learning has been successfully applied to solve the few-shot learning problem, which mainly includes model-based approaches, optimization-based approaches and metric-based approaches. Specifically, for model-based methods, like MANN \cite{DBLP:journals/corr/SantoroBBWL16} and MetaNet \cite{DBLP:conf/icml/MunkhdalaiY17}, they depend on the models which can update the parameters rapidly with a few training steps. For optimization-based methods, like LSTM Meta-Learner \cite{DBLP:conf/iclr/RaviL17}, MAML \cite{DBLP:conf/icml/FinnAL17}, Bayesian MAML \cite{DBLP:conf/nips/YoonKDKBA18}, they intend to adjust some optimization algorithms so that the model can be good at learning with a few examples. For metric-based methods, such as matching network \cite{DBLP:conf/nips/VinyalsBLKW16}, prototypical network \cite{DBLP:conf/nips/SnellSZ17}, relation network \cite{DBLP:conf/cvpr/SungYZXTH18} and so on \cite{DBLP:journals/corr/abs-1902-10482,DBLP:conf/iclr/BaoWCB20}, their basic idea is to learn a feature mapping function that projects support and query samples into an embedding space and classify the queries by learning their relations by some metrics in that space. Among these methods, due to the simplicity and effectiveness, prototypical network is one of the most popular methods in few-shot learning.

\subsection{Multi-label Few-shot Learning}
Traditional few-shot learning focuses on single-label classification task. However, in many real scenarios a sample often has multiple labels, which gives birth to multi-label few-shot learning. As far as we know, only a few works have been done for this task. In computer vision domain, LaSO \cite{DBLP:conf/cvpr/AlfassyKASHFGB19} is a multi-label few-shot image classification model which leverages the label set operations (intersection, union, subtraction) to guide the model to learn the semantic features. In audio domain, \citet{DBLP:conf/mmsp/ChengCY19} use the one-versus-rest episode selection strategy and attention mechanism to deal with the multi-label few-shot sound event recognition problem. In natural language processing domain, \citet{DBLP:conf/aaai/HouLWCL21} focus on multi-label few-shot intent classification task and propose a meta calibrated threshold mechanism with kernel regression and logits adapting that estimates threshold using both prior domain experience and new domain knowledge. 

Proto-AWATT \cite{DBLP:conf/acl/HuZGXGGCS20} is the first work which aims to address aspect category detection in the few-shot scenario. It attempts to leverage support-set and query-set attention mechanisms to alleviate the negative effect caused by noisy aspects, and has achieved the state-of-the-art performance. However, it still suffers from the issue that different aspects perhaps share an identical prototype. In addition, Proto-AWATT learns a dynamic threshold via the policy network, which requires a more complex two-stage training process and that the threshold satisfies the idealized Beta distribution assumption.

\section{Problem Formulation}
To ease understanding, we briefly introduce the task of multi-label few-shot aspect category detection. Table \ref{Tab.1} summarizes some symbol explanation in details.

\emph{\textbf{Few-shot learning}} aims to recognize unknown categories with few labeled samples by leveraging prior knowledge learned from known categories. In general, the data can be divided into two parts: seen (known) class set $\mathcal{C}_\text{seen}$ and unseen (unknown) class set $\mathcal{C}_\text{unseen}$, and $\mathcal{C}_\text{seen} \bigcap \mathcal{C}_\text{unseen}=\emptyset$. A classifier is trained with numerous samples from $\mathcal{C}_\text{seen}$, and it is quickly adopted to $\mathcal{C}_\text{unseen}$ (which is unavailable in training) with only a few labeled data. Meta learning is an effective solution for few-shot learning, which contains two phases: meta-training and meta-testing, and it commonly follows the $N$-way $K$-shot setting, i.e., for each task, there are $N$ classes and each class has $K$ supports (labeled samples).

In meta-training phase, the meta-classifier is trained on $N_{train}$ tasks. In each training task, it consists of a support set and a query set. To construct the training task, $N$ classes are randomly sampled from $\mathcal{N}_\text{seen}$. The support set is composed of randomly selecting $K$ labeled samples from each of the $N$ classes, i.e., $\mathcal{S}=\{(\textbf{\textit{x}}_i, \bm y_i)\}_{i=1}^m$, where $\textbf{\textit{x}}_i$ is a data sample, $\bm y_i$ is the class label and $m=N\times K$. The query set consists of a portion of the remaining samples from these $N$ classes, i.e., $\mathcal{Q}=\{(\textbf{\textit{x}}_j, \bm y_j)\}_{j=1}^n$, where $n$ is the number of queries.

In meta-testing phase, the trained meta-classifier is used to predict the labels of queries in $N_{test}$ tasks. In each test task, it also has a support set and a query set. In a similar manner, $N$ classes are randomly sampled from the test classes $\mathcal{C}_\text{unseen}$. The support set and query set are constructed in the same way as those in meta-training phase. As the labels of queries are unknown in testing stage, the query set in the test task can be represented as $\mathcal{Q}=\{\textbf{\textit{x}}_j\}_{j=1}^n$. The goal is to predict the class labels for these queries.

\emph{\textbf{Multi-label few-shot aspect category detection}} allows that each single sentence is associated with a set of aspect categories simultaneously. Specifically, given a sentence $\textbf{\textit{x}}$, its label can be represented with a vector $\textbf{\textit{y}}=\{y^1,y^2,...,y^N\} \in \mathbb{R}^{N}$, where $y^i\in \{0,1\}$ and $N$ is the number of possible aspects. In this paper, we focus on the multi-label few-shot aspect category detection problem.

\begin{table}[t]
	\caption{Symbol explanation.}
	\small
	\centering
	\linespread{1.8}
	\begin{tabular}{l|l}
		\toprule
		\multicolumn{1}{c|}{Symbol} & \multicolumn{1}{c}{Explanation}  \\
		\midrule
		$\mathcal{C}_\text{seen}$       & the seen class set  \\
		$\mathcal{C}_\text{unseen}$     & the unseen class set\\
		$ N $ 							& the number of aspect classes in each episode  \\
		$ K $ 							& the number of support shots in each class  \\
		$ \mathcal{S} $					& the support set of an episode				\\
		$ \mathcal{Q} $					& the query set of an episode                                \\
		$ \textbf{\textit{x}} $         & a sentence with $T$ words\\
		$ \textbf{\textit{y}} $         & the class label of $ \textbf{\textit{x}} $\\
		$\bm H$				            & the embedding matrix of $ \textbf{\textit{x}} $ via any pre-trained model\\
		$\bm o$                         & the representation of $ \textbf{\textit{x}} $  via feature extraction\\
		$\bm E$                         & the label description representation\\
		$\bm p^i$                       & the label-enhanced prototype of class $i$\\
		$\bm z^i$                       & the label-specific embedding associated with class $i$\\
		\bottomrule
	\end{tabular}
	\label{Tab.1}
\end{table}

\begin{figure*}[t] 
	\centering
	\includegraphics[width=0.9\linewidth]{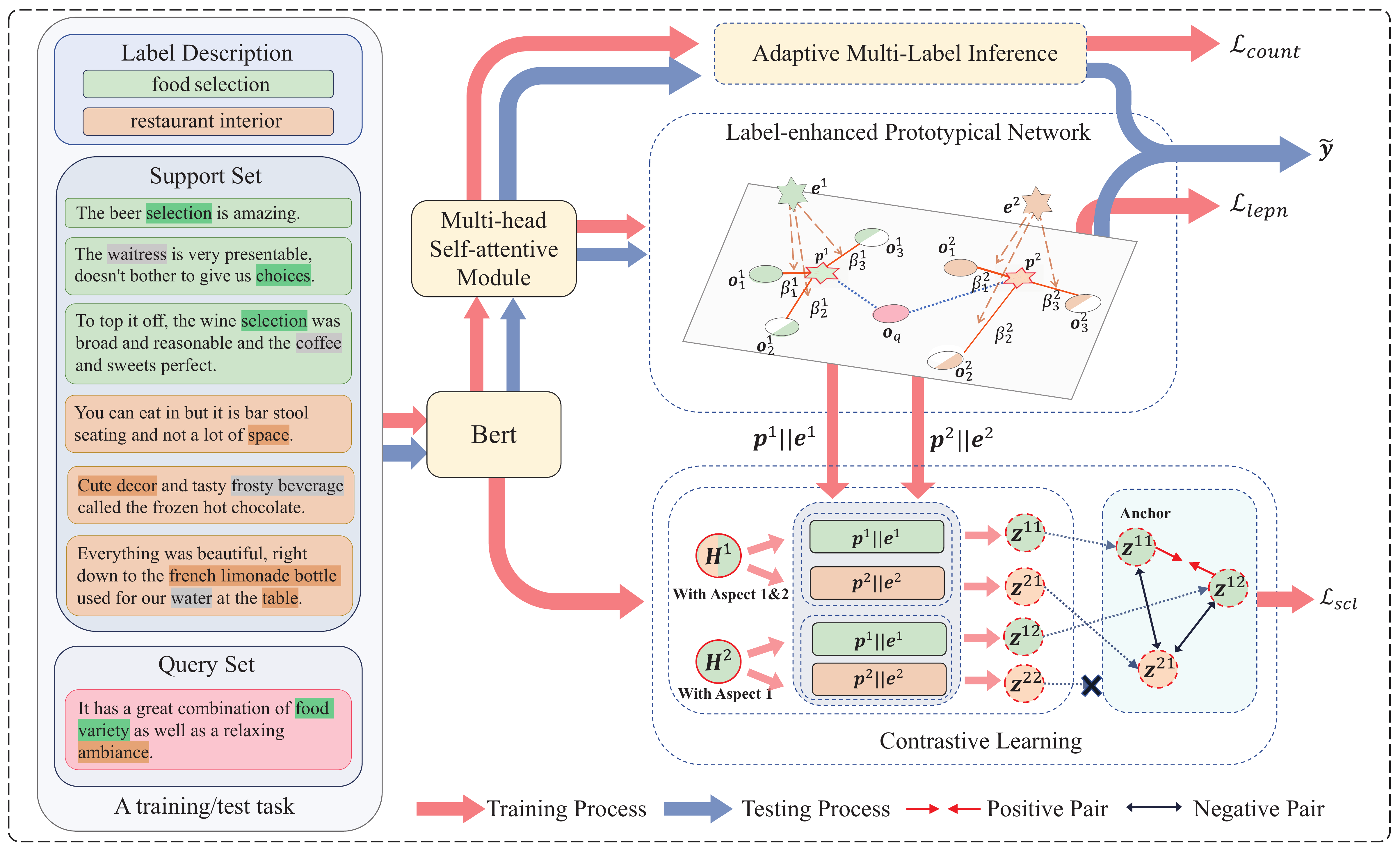}
	\caption{Illustration of our proposed method LPN.} 
	\label{framework} 
\end{figure*}

\section{Approach}
The overall framework of the proposed LPN is shown in Figure \ref{framework}. It consists of four components: feature extraction, label-enhanced prototypical network, contrastive learning and adaptive multi-label inference. In this section, we will introduce these modules in details.

\subsection{Feature Extraction}
Given a sentence $\textbf{\textit{x}}$ with $T$ words, we can use any pre-trained language model like Bert \cite{DBLP:conf/naacl/DevlinCLT19} to encode each word (token), and then get the embedding matrix $\textbf{\textit{H}} = [ \textbf{\textit{h}}_1, \textbf{\textit{h}}_2, ... ,\textbf{\textit{h}}_T ] \in \mathbb{R} ^{d \times T}$. To better extract the sentence-level semantic feature and assign reasonable importance for each word, we follow \cite{DBLP:conf/iclr/LinFSYXZB17,DBLP:conf/acl/YanFLLZWL20} to utilize a multi-head self-attentive module to generate the sentence embedding. Specifically, 
\begin{equation}\label{self_attentive}
    \bm A =\text{softmax}\left( \bm F_2 \text{tanh}(\bm F_1 \bm H)\right), 
\end{equation}
where $\bm A \in \mathbb{R}^{R \times T}$ is the self-attention weight matrix, $R$ is the number of independent attention heads. $\bm F_{1} \in \mathbb{R}^{d' \times d}$ and $\bm F_2 \in \mathbb{R}^{R \times d'}$ are trainable parameter matrices. After obtaining $\bm A$, we first calculate the embedding matrix by:
\begin{equation}\label{M_calculate}
\bm M = \bm H \bm A^T,
\end{equation}
where $\bm M =[\bm m_1, \bm m_2, ..., \bm m_R] \in \mathbb{R}^{d \times R}$. Furthermore, we concatenate the obtained embeddings from different heads, and use a simple linear projection to calculate the embedding of the sentence, 
\begin{equation}\label{sentence_emb}
\bm o = \bm F_3  [\bm m_1|| \bm m_2 || ... ||\bm m_R ],
\end{equation}
where $\bm F_3\in\mathbb{R}^{d\times dR}$ is a trainable parameter matrix, and $||$ represents the concatenation operation. $\bm o \in \mathbb{R}^d$ is the final representation of the sentence $\textbf{\textit{x}}$.

\subsection{Label-enhanced Prototypical Network}
In terms of multi-label few-shot aspect category detection task, the key point is to learn more discriminative prototypes which could better retain class-relevant information while eliminating the harmful effect caused by other noisy (irrelevant) aspects. To achieve this goal, we propose to leverage the label text description information to calculate the relationship between sentences and aspect labels, thus obtaining more representative prototypes.

Considering the $N$-way $K$-shot setting, we have a support set $\mathcal{S}$, which can be represented by $\mathcal{S}=\{\textbf{\textit{x}}_1^1, \textbf{\textit{x}}_2^1, ... ,\textbf{\textit{x}}_K^1, ... ,\textbf{\textit{x}}_1^N, \textbf{\textit{x}}_2^N, ... ,\textbf{\textit{x}}_K^N\}$, where $\textbf{\textit{x}}_j^i$ denotes the $j$-th sample belonging to the $i$-th class. After feature extraction, we get the representations of these samples $\mathcal{O}=\{\bm o_1^1, \bm o_2^1, ... ,\bm o_K^1, ... ,\bm o_1^N, \bm o_2^N, ... ,\bm o_K^N\}$. In a similar manner, for each label description like "Room cleanliness" or "Staff owner", we can obtain its corresponding representation via feature extraction. In the $N$-way scenario, the label description representation can be represented as $\bm E=\{\bm e^1,\bm e^2, ... ,\bm e^N\}$. 

When calculating the class prototype, as each sentence may contain multiple aspects, we would better first determinate the importance weight of each sentence for a class prototype. To this end, we utilize the label text description as auxiliary information. Specifically, 
\begin{equation}\label{alpha}
	\alpha_j^i = {\textbf{\textit{o}}_j^i}^T \bm W \textbf{\textit{e}}^i,
\end{equation}
where $\alpha_j^i$ denotes the importance weight of the $j$-th sentence for the $i$-th class prototype. $\textbf{\textit{o}}_j^i  \in \mathbb{R}^{d}$ denotes the representation of the $j$-th sample belonging to the $i$-th class, ${\textbf{\textit{o}}_j^i}^T \in \mathbb{R}^{1 \times d}$ is the transpose of $\textbf{\textit{o}}_j^i$.
$\bm W \in \mathbb{R}^{d \times d}$ is a trainable projection matrix. $\bm e^i$ is the representation of the $i$-th label description. 

Inspired by low-rank bilinear model \cite{DBLP:journals/tnn/YuYXFT18}, if imposing a low-rank restriction on $\bm W$, Eq.~(\ref{alpha}) can be rewritten as follows:
\begin{equation}\label{alpha_improve}
\begin{split}
	\alpha_j^i &= {\textbf{\textit{o}}_j^i}^T \bm U \bm V^T \textbf{\textit{e}}^i = \bm 1^T (\bm U^T \textbf{\textit{o}}_j^i \circ  \bm V^T\textbf{\textit{e}}^i  ),
\end{split}
\end{equation}
where $\bm U \in \mathbb{R}^{d \times k}$ and $\bm V \in \mathbb{R}^{d \times k}$ are two low-rank matrices with $k<d$. $\bm 1^T$ is a all-one vector. $\circ$ is the Hadamard product, i.e., element-wise multiplication. By using this low-rank trick, we can reduce the number of parameters to some extent.

To make the coefficients comparable among different sentences, we normalize them across $K$ sentences (shots) belonging to the same class with the softmax function:
\begin{equation}
	\beta_j^i = \frac{\text{exp}(\alpha_j^i)}{\sum^{K}_{j'=1} \text{exp}({{\alpha_{j'}^{i}}})}.
\end{equation}

Then we calculate the label-enhanced prototype $\textbf{\textit{p}}^i \in \mathbb{R}^{d}$ for class $i$ by:
\begin{equation}
	\textbf{\textit{p}}^i = \sum_{j=1}^{K}\beta_j^i\textbf{\textit{o}}_j^i.
\end{equation}

Given a query sentence $\textbf{\textit{x}} \in \mathcal{Q}$, we can compute the conditional probability $p(y=y^i|\textbf{\textit{x}}, \mathcal{S})$ to predict its aspect label based on negative squared Euclidean distance.
\begin{equation}
p(y=y^i|\textbf{\textit{x}}, \mathcal{S}) = \frac{\text{exp}(-|| \bm o - \textbf{\textit{p}}^i ||_2^2)}{\sum_{j=1}^{N}\text{exp}(-|| \bm o - \textbf{\textit{p}}^{j} ||_2^2)},
\end{equation}
where $\bm o$ denotes the representation of $\textbf{\textit{x}}$, which is obtained via feature extraction.

Finally, we perform the cross-entropy loss on all samples in the query set $\mathcal{Q}$, i.e., the loss function of label-enhanced prototypical network $\mathcal{L}_{lepn}$ can be written as:
\begin{equation}\label{lepn}
	\mathcal{L}_{lepn} = \frac{1}{\left|\mathcal{Q}\right|} \sum_{\textbf{\textit{x}}\in \mathcal{Q}} \sum_{i=1}^{N}{ - y^i \text{log} p (y=y^i|\textbf{\textit{x}}, \mathcal{S})},
\end{equation}
where $\left|\mathcal{Q}\right|$ is the number of samples in  $\mathcal{Q}$. Note that in the multi-label $N$-way $K$-shot setting, as a sentence may have multiple labels, we need to consider $N$ labels for each query sentence.

\subsection{Integrating with Contrastive Learning}
Contrastive learning has achieved great success in computer vision \cite{DBLP:conf/cvpr/00230W0W21, DBLP:conf/cvpr/ChenWLDB21}, which aims to maximize similarities between instances from the same class and minimize similarities between instances from different classes. Here we integrate the contrastive learning into label-enhanced prototypical network to generate better sentence embeddings.

For traditional single-label aspect detection, we can directly construct the contrastive samples using the known aspect labels. However, in the multi-label aspect detection scenario, as a sentence may contain a couple of aspects, for example, one sentence "The pool is gorgeous, the room clean, delicious food, and staff that went above and beyond to help us enjoy our stay" contains two aspect labels "food" and "staff", and the other sentence "The food is always good and service has always been a great experience" contains two aspect labels "food" and "experience". If simply treating these two sentences as positive pairs, it is unreasonable obviously. The reason is that though these two sentences share a common aspect label "food", they also have a totally different aspect label. To alleviate the above issue, we use the prototypes and label description information to first generate the label-specific embeddings for each sentence, and then construct the contrastive samples.

In the $N$-way $K$-shot setting, for each meta-task, we can obtain $N$ prototypes $\bm P = \{\textbf{\textit{p}}^1, \textbf{\textit{p}}^2, ..., \textbf{\textit{p}}^N \}$ and $N$ label description representations $\bm E = \{\textbf{\textit{e}}^1, \textbf{\textit{e}}^2, ..., \textbf{\textit{e}}^N\}$. By combining $\bm P$ and $\bm E$, we can have the prototypes integrated with label description information $\{\textbf{\textit{a}}^1, \textbf{\textit{a}}^2, ..., \textbf{\textit{a}}^N\}$, where $\textbf{\textit{a}}^i = [\textbf{\textit{p}}^i || \textbf{\textit{e}}^i] \in \mathbb{R}^{2d}$ and $||$ represents the concatenation operation. Then for a sentence $\textbf{\textit{x}}$ in the meta-task, we can compute its label-specific embedding $\textbf{\textit{z}}^i \in \mathbb{R}^d$ associated with label $i$ by:
\begin{equation}\label{10}
	\textbf{\textit{z}}^i = \bm g^i \bm H^T,
\end{equation}
where $\textbf{\textit{H}} = [ \textbf{\textit{h}}_1, \textbf{\textit{h}}_2, ... ,\textbf{\textit{h}}_T ] \in \mathbb{R} ^{d \times T}$ is the embedding matrix obtain from any pre-trained language model like Bert. $\bm g^ i\in \mathbb{R}^{1 \times T}$ is a weight vector obtained by:
\begin{equation}\label{11}
	\bm g^i = \text{softmax}((\bm W_a \bm a^i + \bm b_a)^T \bm H),
\end{equation}
where $\bm W_a \in \mathbb{R}^{d \times 2d}$ and $\bm b_a \in \mathbb{R}^d$ are trainable parameters.

In meta-training phase, given a meta-task with $N_t$ samples, we first use Eq.~(\ref{10}) and (\ref{11}) to get the label-specific embeddings and then collect all these embeddings to construct the set $\bm Z = \{\textbf{\textit{z}}^{ij} \in \mathbb{R}^{d}|i \in \{1,2,...,N\}, j \in \{1,2,...,N_t\}\}$. If we regard each label-specific embedding as an independent instance, each $\textbf{\textit{z}}^{ij}$ will be associated with a single ground-truth label $y^{ij}$. Specifically, a sentence "The food is always good and service has always been a great experience" contains two aspect labels "food" and "experience", the labels of obtained label-specific embedding associated with "food" and "experience" will be set to 1. Then we can define the set $\bm Y = \{y^{ij} \in \{0,1\}|i \in \{1,2,...,N\}, j \in \{1,2,...,N_t\}\}$. Furthermore, we define the set $\bm I=\{\textbf{\textit{z}}^{ij} \in \bm Z| y^{ij}=1\}$ which contains the label-specific embeddings with usable ground-truth labels, and the set $\bm \Gamma^{ij}=\{ \bm I \backslash \textbf{\textit{z}}^{ij} \}$ which contains the embeddings in $\bm I$ with $\textbf{\textit{z}}^{ij}$  excluded. Considering $\textbf{\textit{z}}^{ij}$ as the anchor, we can generate the positive sample set $\bm \Lambda^{ij}= \{\textbf{\textit{z}}^{ik} \in \bm \Gamma^{ij} | y^{ik} = y^{ij} =1\}$ for $\textbf{\textit{z}}^{ij}$, and the negative samples for $\textbf{\textit{z}}^{ij}$ are the remaining ones in $\bm \Gamma^{ij}$. With above notations, the contrastive learning loss for the anchor $\textbf{\textit{z}}^{ij}$ can be written as:
\begin{equation}\label{eq.scl1}
	\mathcal{L}_{scl}^{ij} = -\frac{1}{|\bm \Lambda^{ij}|}  \sum_{\textbf{\textit{z}}^{ik} \in \bm \Lambda^{ij}} \text{log}\frac{\text{exp}(\textbf{\textit{z}}^{ij} \cdot \textbf{\textit{z}}^{ik} / \tau )}{\sum_{\bm z^*\in \bm \Gamma^{ij}}  \text{exp}(\textbf{\textit{z}}^{ij} \cdot \textbf{\textit{z}}^* / \tau ) },
\end{equation}
where $\textbf{\textit{z}}^{ij} \cdot \textbf{\textit{z}}^{ik}$ denotes the inner product of the two vectors. $|\bm \Lambda^{ij}|$ is the number of embeddings in $\bm \Lambda^{ij}$. $\tau > 0$ is an adjustable scalar parameter, which can control the separation degree of classes \cite{DBLP:conf/iclr/GunelDCS21}. By considering all the anchors, we can have the entire contrastive loss as follows.
\begin{equation}\label{eq.scl2}
	\mathcal{L}_{scl} = \frac{1}{|\bm I|}  \sum_{\textbf{\textit{z}}^{ij} \in \bm I} \mathcal{L}_{scl}^{ij}.
\end{equation}

To analyze Eq.~(\ref{eq.scl2}), we do some simple formula transformation as below.
\begin{equation}
	\begin{split}
		\mathcal{L}_{scl} &  =\frac{1}{|\bm I|}  \sum_{\textbf{\textit{z}}^{ij} \in \bm I} -\frac{1}{|\bm \Lambda^{ij}|}  \mathcal{L'}, \\
		 \mathcal{L'} & =\sum_{\textbf{\textit{z}}^{ik} \in \bm \Lambda^{ij}} \text{log}\frac{\text{exp}(\textbf{\textit{z}}^{ij} \cdot \textbf{\textit{z}}^{ik} / \tau )}{\sum_{z^*\in \bm \Gamma^{ij}}  \text{exp}(\textbf{\textit{z}}^{ij} \cdot \textbf{\textit{z}}^* / \tau ) } \\
		 &=\sum_{\textbf{\textit{z}}^{ik} \in \bm \Lambda^{ij}} (\underbrace{\frac{\textit{\textbf{z}}^{ij}\cdot\textit{\textbf{z}}^{ik}}{\tau}}_{positive} -\text{log}  \underbrace{\sum_{\bm z^*\in \bm \Gamma^{ij}}{\text{exp}(\frac{\textit{\textbf{z}}^{ij}\cdot\textit{\textbf{z}}^*}{\tau})}}_{positive+negative}).
	\end{split}		
\end{equation}

From the above formula, it is easy to find that if we want to minimize $ \mathcal{L}_{scl} $, we must maximize $ \mathcal{L'}$, where we need to maximize the positive term and minimize the positive+negative term, so the negative term will be decreased. Intuitively, the contrastive learning technique can push the label-specific embeddings from the same class close and embeddings from different classes further apart.

\subsection{Adaptive Multi-label Inference}
For multi-label few-shot aspect category detection, one of the challenges is to determine the number of aspects in the sentence. Previous work \cite{DBLP:conf/acl/HuZGXGGCS20} learns a dynamic threshold via the policy network. However, it requires that the threshold satisfies the Beta distribution assumption, which seems a little over-idealized. In addition, as it is a two-stage method, the training process is also more complicated. To overcome this issue, we propose an adaptive multi-label inference method, which is simple yet effective.

In the $N$-way $K$-shot setting, $N$ is the maximal number of aspects in a sentence. Given a sentence $\textit{\textbf{x}}$, we can get its representation $\bm o \in \mathbb{R}^d$ via feature extraction. Then we use a multi-layer perception to predict the number of aspects in $\textit{\textbf{x}}$. Specifically,
\begin{equation}
	\textit{\textbf{n}}_l = \text{softmax}( \bm W_l \textit{\textbf{o}} + \bm b_l),
\end{equation}
where $\bm W_l  \in \mathbb{R}^{N \times d}$ and $\bm b_l \in \mathbb{R}^N$ are trainable parameters. $\textit{\textbf{n}}_l \in \mathbb{R}^N$ is the indicator for the number of aspects. Take an example, if the maximal value of $\textit{\textbf{n}}_l$ is the second element, it means that $\textit{\textbf{x}}$ contains two aspects. 

Then in the meta-training stage, for each sentence from support set $\mathcal{S}$ and query set $\mathcal{Q}$, we use cross entropy to calculate the loss of the aspect count,
\begin{equation}\label{count}
	\mathcal{L}_{count} = \frac{1}{\left|\mathcal{S} \cup \mathcal{Q}\right|} \sum_{\textit{\textbf{x}} \in \mathcal{S} \cup \mathcal{Q}} - \bm{1}^T (\textit{\textbf{t}}_l \circ  \text{log}(\textit{\textbf{n}}_l)),
\end{equation}
where $\bm 1^T$ is a all-one vector. $\circ$ is the Hadamard product, i.e., element-wise multiplication. $\text{log}(\textit{\textbf{n}}_l) \in \mathbb{R}^N$ is to do the log operation on each element of $\textit{\textbf{n}}_l$. $\textit{\textbf{t}}_l$ is the ground-truth aspect count vector of $\textit{\textbf{x}}$.

By combining Eq.~(\ref{lepn}), (\ref{eq.scl2}) and (\ref{count}), we have the overall loss function of the proposed framework:
\begin{equation}
	\mathcal{L}_{total} = \mathcal{L}_{lepn} + \gamma \mathcal{L}_{scl} + \lambda \mathcal{L}_{count},
\end{equation}
where $\gamma$ and $\lambda$ are adjustable trade-off parameters. By minimizing $\mathcal{L}_{total}$ with the gradient descent method, all trainable parameters can be learned.

\section{Experiments}

\subsection{Datasets}

\begin{table}[t]
	\caption{Dataset statistics. \#Aspects and \#Sentences denote the number of aspects and sentences respectively.}
	\begin{tabular}{llcc}
		\toprule[1pt]
		Dataset                         & Split     & \#Aspects   & \#Sentences \\ \midrule
		\multirow{3}{*}{FewAsp (single)} & Training & 64       & 12800   \\
		& Validation   & 16      & 3200    \\
		& Testing  & 20       & 4000    \\ \midrule
		\multirow{3}{*}{FewAsp (multi)}  & Training & 64   & 25600   \\
		& Validation   & 16    & 6400    \\
		& Testing  & 20    & 8000    \\ \midrule
		\multirow{3}{*}{FewAsp}         & Training & 64  & 40320   \\
		& Validation   & 16    & 10080   \\
		& Testing  & 20    & 12600   \\ \toprule[1pt]
	\end{tabular}
	\label{tab.1}
\end{table}

For fair comparison, we exactly follow \cite{DBLP:conf/acl/HuZGXGGCS20} to perform experiments on three datasets: FewAsp (single), FewAsp (multi) and FewAsp. All these datasets are sampled from the large-scale multi-domain dataset for aspect recommendation YelpAspect \cite{DBLP:conf/kdd/Bauman0T17}. Specifically, FewAsp (single) consists of singe-aspect sentences, FewAsp (multi) consists of a majority of multi-aspect sentences with a minority of single-aspect sentences, as some aspects only have a small amount of multi-aspect samples, and FewAsp is randomly sampled from the original dataset, which follows the same data distribution with the real scenario. For data split, we also follow \cite{DBLP:conf/acl/HuZGXGGCS20} to divide the 100 aspects without intersection into 64 aspects for training, 16 aspects for validation, and 20 aspects for testing. The detailed dataset statistics is shown in Table \ref{tab.1}.

\begin{table}[t]
	\caption{Hyperparameters of our proposed method LPN.}
	\centering
	\begin{tabular}{cccccccc}
		\toprule
		Model   & $d$ & $d'$ & $R$ & $k$ & $\lambda$ & $\gamma$ &  $\tau$\\
		\midrule
		LPN  & 768 & 256 & 4 & 100 & 0.1 & 0.01 & 0.1\\
		\bottomrule
	\end{tabular}
	\label{parameter}
\end{table}

\begin{table*}[t]
	\caption{Average AUC and macro-F1 score on FewAsp (single).}
	\setlength{\tabcolsep}{3mm}
	\begin{tabular}{l|cc|cc|cc|cc}
			\toprule[1pt]
			\multicolumn{1}{c|}{}                         & \multicolumn{2}{c|}{5-way 5-shot}                                   & 
			\multicolumn{2}{c|}{5-way 10-shot} & \multicolumn{2}{c|}{10-way 5-shot} & \multicolumn{2}{c}{10-way 10-shot} \\
			\multicolumn{1}{c|}{\multirow{-2}{*}{Model}} & AUC             & F1 & AUC              & F1              & AUC              & F1              & AUC               & F1              \\ 
			\midrule
			Matching Network \cite{DBLP:conf/nips/VinyalsBLKW16}  & 97.05          & 81.89    & 97.49           & 84.62           & 96.30           & 70.95           & 96.72            & 73.28           \\
			Prototypical Network \cite{DBLP:conf/nips/SnellSZ17}  & 96.49          & 83.30    & 97.53           & 86.29           & 95.97           & 74.23           & 96.71            & 76.83           \\
			Relation Network \cite{DBLP:conf/cvpr/SungYZXTH18}  & 93.31          & 75.79      & 90.86           & 72.02           & 91.81           & 63.78           & 90.54            & 61.15           \\
			Graph Network \cite{DBLP:conf/iclr/SatorrasE18}  & 96.54          & 81.45         & 97.46           & 85.04           & 95.45           & 70.75           & 96.97            & 77.84           \\
			IMP \cite{DBLP:conf/icml/AllenSST19}     & 96.65          & 83.69                 & 97.47           & 86.14           & 96.00           & 73.80           & 96.91            & 77.09           \\
			Proto-HATT \cite{DBLP:conf/aaai/GaoH0S19}   & 96.45          & 83.33              & 97.62           & 86.71           & 95.71           & 73.42           & 97.00            & 77.65           \\
			Proto-AWATT \cite{DBLP:conf/acl/HuZGXGGCS20}  & 97.56          & 86.71            & 97.96           & 88.54           & 97.01           & 80.28           & 97.55            & 82.97           \\ 
			\midrule
			LPN (o, o)                                     & 97.88          & 87.62           & 98.48           & 90.31           & 98.13           & 83.99           & 98.53            & 85.95           \\
			LPN (w, o)                                     & 99.22          & 92.61           & 99.35           & 93.57           & 99.11           & 89.35           & \textbf{99.32}   & \textbf{91.08}  \\
			LPN (w, w)                                  & \textbf{99.29} & \textbf{94.43}      & \textbf{99.49}  & \textbf{94.40}  & \textbf{99.14}  & \textbf{89.40}  & 99.28            & 90.43           \\
			\bottomrule
	\end{tabular}
	\label{tab.single}
\end{table*}
	
\begin{table*}[t]
	\caption{Average AUC and macro-F1 score on FewAsp (multi).}
	\setlength{\tabcolsep}{3mm}
	\begin{tabular}{l|cc|cc|cc|cc}
			\toprule[1pt]
			\multicolumn{1}{c|}{}                         & \multicolumn{2}{c|}{5-way 5-shot}     & \multicolumn{2}{c|}{5-way 10-shot}                & \multicolumn{2}{c|}{10-way 5-shot}                & \multicolumn{2}{c}{10-way 10-shot}               \\
			\multicolumn{1}{c|}{\multirow{-2}{*}{Model}} & \multicolumn{1}{c}{AUC} & \multicolumn{1}{c|}{F1} & \multicolumn{1}{c}{AUC} & \multicolumn{1}{c|}{F1} & \multicolumn{1}{c}{AUC} & \multicolumn{1}{c|}{F1} & \multicolumn{1}{c}{AUC} & \multicolumn{1}{c}{F1} \\ 
			\midrule
			Matching Network \cite{DBLP:conf/nips/VinyalsBLKW16} & 89.54                  & 65.70   & 91.38                  & 69.02                   & 88.28                  & 50.86                   & 89.94                  & 54.42                   \\
			Prototypical Network \cite{DBLP:conf/nips/SnellSZ17} & 89.67                  & 67.88   & 91.60                   & 72.32                   & 88.01                  & 52.72                   & 90.68                  & 58.92                   \\
			Relation Network \cite{DBLP:conf/cvpr/SungYZXTH18}   & 84.91                  & 58.38   & 86.21                  & 61.37                   & 84.22                  & 43.71                   & 84.72                  & 44.85                   \\
			Graph Network \cite{DBLP:conf/iclr/SatorrasE18}   & 87.97                  & 59.25      & 90.45                  & 64.63                   & 86.05                  & 45.42                   & 88.44                  & 48.49                   \\
			IMP  \cite{DBLP:conf/icml/AllenSST19}    & 90.12                  & 68.86               & 92.29                  & 73.51                   & 88.71                  & 53.96                   & 91.10                  & 59.86                   \\
			Proto-HATT \cite{DBLP:conf/aaai/GaoH0S19}   & 91.10                  & 69.15            & 93.03                  & 73.91                   & 90.44                  & 55.34                   & 92.38                  & 60.21                   \\
			Proto-AWATT \cite{DBLP:conf/acl/HuZGXGGCS20}    & 91.45                  & 71.72        & 93.89                  & 77.19                   & 89.80                  & 58.89                   & 92.34                  & 66.76                   \\ 
			\midrule
			LPN (o, o)                                    & 93.09                  & 72.45          & 94.92                  & 76.89                   & 92.95                  & 61.33                   & 94.62                  & 66.39                   \\
			LPN (w, o)                                     & 95.43                  & 78.82         & 96.22                  & 81.70                   & 94.29                  & 66.36                   & 95.43               & 71.08               \\
			LPN (w, w)                                  & \textbf{95.66}         & \textbf{79.48}    & \textbf{96.55}         & \textbf{82.81}          & \textbf{94.51}         & \textbf{67.28}          & \textbf{95.66}         & \textbf{71.87}      \\ 
			\bottomrule
	\end{tabular}
	\label{tab.multi}
\end{table*}
	
\begin{table*}[t]
\caption{Average AUC and macro-F1 score on FewAsp.}
\setlength{\tabcolsep}{3mm}
	\begin{tabular}{l|cc|cc|cc|cc}
	\toprule[1pt]
			\multicolumn{1}{c|}{}                         & \multicolumn{2}{c|}{5-way 5-shot}       & \multicolumn{2}{c|}{5-way 10-shot}                & \multicolumn{2}{c|}{10-way 5-shot}                & \multicolumn{2}{c}{10-way 10-shot}               \\
			\multicolumn{1}{c|}{\multirow{-2}{*}{Model}} & \multicolumn{1}{c}{AUC} & \multicolumn{1}{c|}{ F1} & \multicolumn{1}{c}{AUC} & \multicolumn{1}{c|}{F1} & \multicolumn{1}{c}{AUC} & \multicolumn{1}{c|}{F1} & \multicolumn{1}{c}{AUC} & \multicolumn{1}{c}{F1} \\ 
			\midrule
			Matching Network \cite{DBLP:conf/nips/VinyalsBLKW16} & 90.76                  & 67.14   & 92.39                  & 70.09                   & 88.44                  & 51.27                   & 89.90                  & 54.61                   \\
			Prototypical Network \cite{DBLP:conf/nips/SnellSZ17} & 88.88                  & 66.96   & 91.77                  & 73.27                   & 87.35                  & 52.06                   & 90.13                  & 59.03                   \\
			Relation Network \cite{DBLP:conf/cvpr/SungYZXTH18}  & 85.56                  & 59.52    & 86.98                  & 62.78                   & 84.94                  & 45.62                   & 83.77                  & 44.70                    \\
			Graph Network \cite{DBLP:conf/iclr/SatorrasE18} & 89.48                  & 61.49        & 92.35                  & 69.89                   & 87.35                  & 47.91                   & 90.19                  & 56.06                   \\
			IMP \cite{DBLP:conf/icml/AllenSST19}   & 89.95                  & 68.96                 & 92.30                   & 74.13                   & 88.50                  & 54.14                   & 90.81                  & 59.84                   \\
			Proto-HATT \cite{DBLP:conf/aaai/GaoH0S19}      & 91.54                  & 70.26         & 93.43                  & 75.24                   & 90.63                  & 57.26                   & 92.86                  & 61.51                   \\
			Proto-AWATT \cite{DBLP:conf/acl/HuZGXGGCS20}        & 93.35                  & 75.37    & 95.28                  & 80.16                   & 92.06                  & 65.65                   & 93.42                  & 69.70                   \\ 
			\midrule
			LPN (o, o)                                     & 94.15                  & 76.19         & 95.85                  & 80.37                   & 94.03                  & 65.72                   & 94.98                  & 69.22                   \\
			LPN (w, o)                                     & 96.41                  & \textbf{82.26}   & \textbf{97.43}         & \textbf{85.81}          & 95.26                  & 71.25                   & 96.23                  & 75.49                   \\
			LPN (w, w)                                  & \textbf{96.45}         & 82.22               & 97.15                  & 84.90                    & \textbf{95.36}         & \textbf{71.42}          & \textbf{96.55}         & \textbf{76.51}          \\
	\bottomrule
	\end{tabular}
	\label{tab.all}
\end{table*}
	
\subsection{Baselines}
We compare the proposed LPN model with the following strong baselines: Matching Network \cite{DBLP:conf/nips/VinyalsBLKW16}, Relation Network \cite{DBLP:conf/cvpr/SungYZXTH18}, Graph Network \cite{DBLP:conf/iclr/SatorrasE18}, Prototypical Network \cite{DBLP:conf/nips/SnellSZ17}, IMP \cite{DBLP:conf/icml/AllenSST19}, Proto-HATT \cite{DBLP:conf/aaai/GaoH0S19} and Proto-AWATT \cite{DBLP:conf/acl/HuZGXGGCS20}.

\begin{itemize}
	\item \textbf{Matching Network} \cite{DBLP:conf/nips/VinyalsBLKW16} first learns a embedding mapping function and then takes the cosine similarity as distance measure to obtain the classification results.
	\item \textbf{Prototypical Network} \cite{DBLP:conf/nips/SnellSZ17} calculates the prototype for each class by averaging the corresponding support samples, and utilizes the negative Euclidean distance between query samples and prototypes to do the few-shot classification task.
	\item \textbf{Relation Network} \cite{DBLP:conf/cvpr/SungYZXTH18} uses a deep neural network instead of the fixed distance measure to calculate the relationship between query and support samples.
	\item \textbf{Graph Network} \cite{DBLP:conf/iclr/SatorrasE18} attempts to cast few-shot learning as a supervised message passing task which is trained end-to-end using graph neural networks.
	\item \textbf{IMP} \cite{DBLP:conf/icml/AllenSST19} introduces infinite mixture prototypes for few-shot learning, which represents each class by a set of clusters.
	\item \textbf{Proto-HATT} \cite{DBLP:conf/aaai/GaoH0S19} is a hybrid attention-based prototypical networks for the problem of noisy few-shot relation classification. It uses instance-level and feature-level attention schemes to highlight the crucial instances and features respectively.
	\item \textbf{Proto-AWATT} \cite{DBLP:conf/acl/HuZGXGGCS20} is the first method for multi-label few-shot aspect category detection task. It utilizes support-set and query-set attention mechanisms to alleviate the adverse effect caused by noisy aspects.
\end{itemize}

In addition, we conduct ablation study to evaluate the contribution of label enhancement and contrastive learning in LPN. Specifically, we evaluate LPN under three cases: LPN (o, o), LPN (w, o) and LPN (w, w). 

\begin{itemize}
	\item \textbf{LPN (o, o)} means the LPN model without label enhancement and contrastive learning.
	\item \textbf{LPN (w, o)} means the LPN model with label enhancement, but without contrastive learning.
	\item \textbf{LPN (w, w)} means the LPN model with label enhancement and contrastive learning, i.e., the final model.
\end{itemize}

\begin{figure}[t]
		\centering
		\subfigure[AUC]{
			\includegraphics[height=3.5cm,width=4cm]{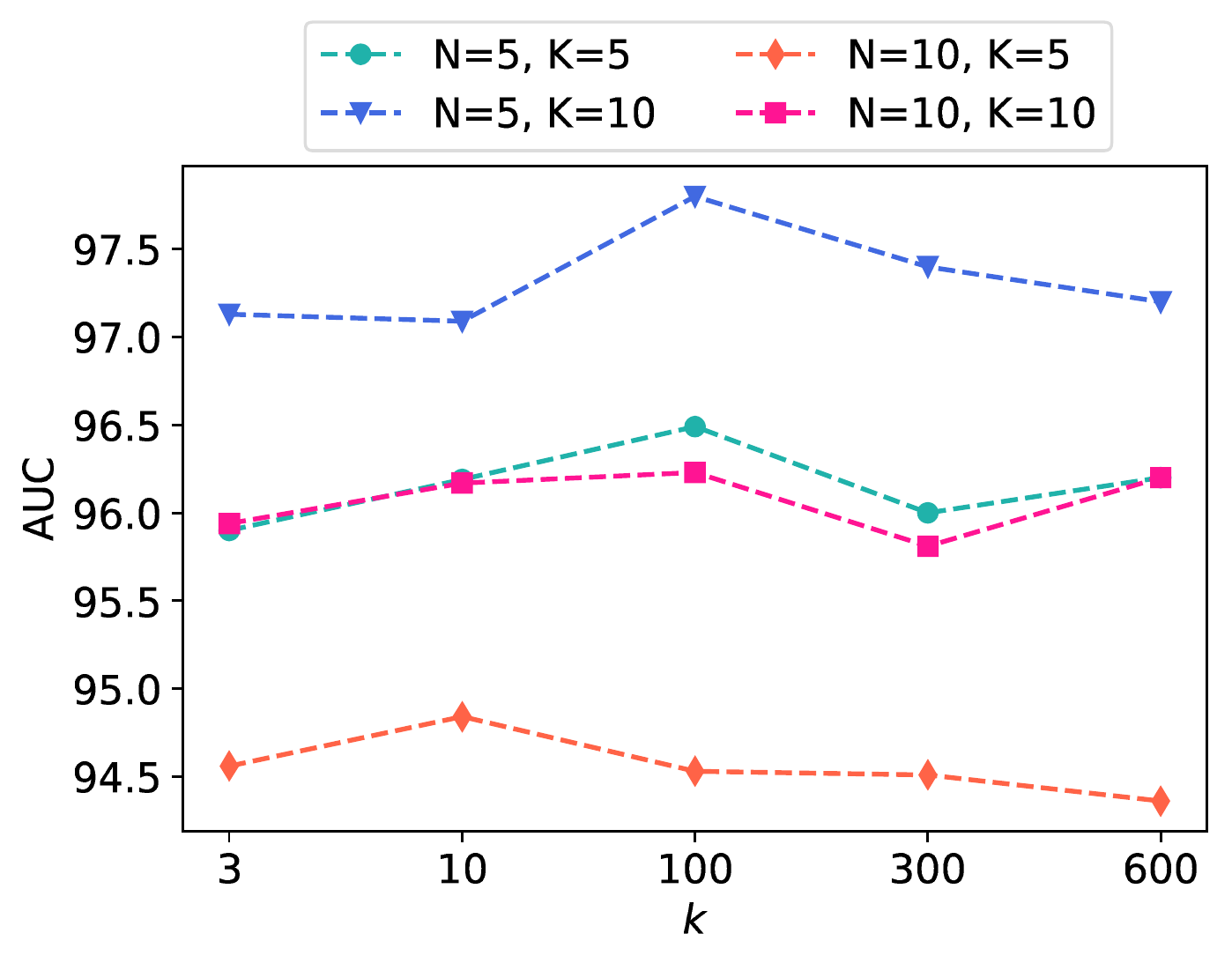}
			\label{ablation_f}}
		\subfigure[F1 Score]{
			\includegraphics[height=3.5cm,width=4cm]{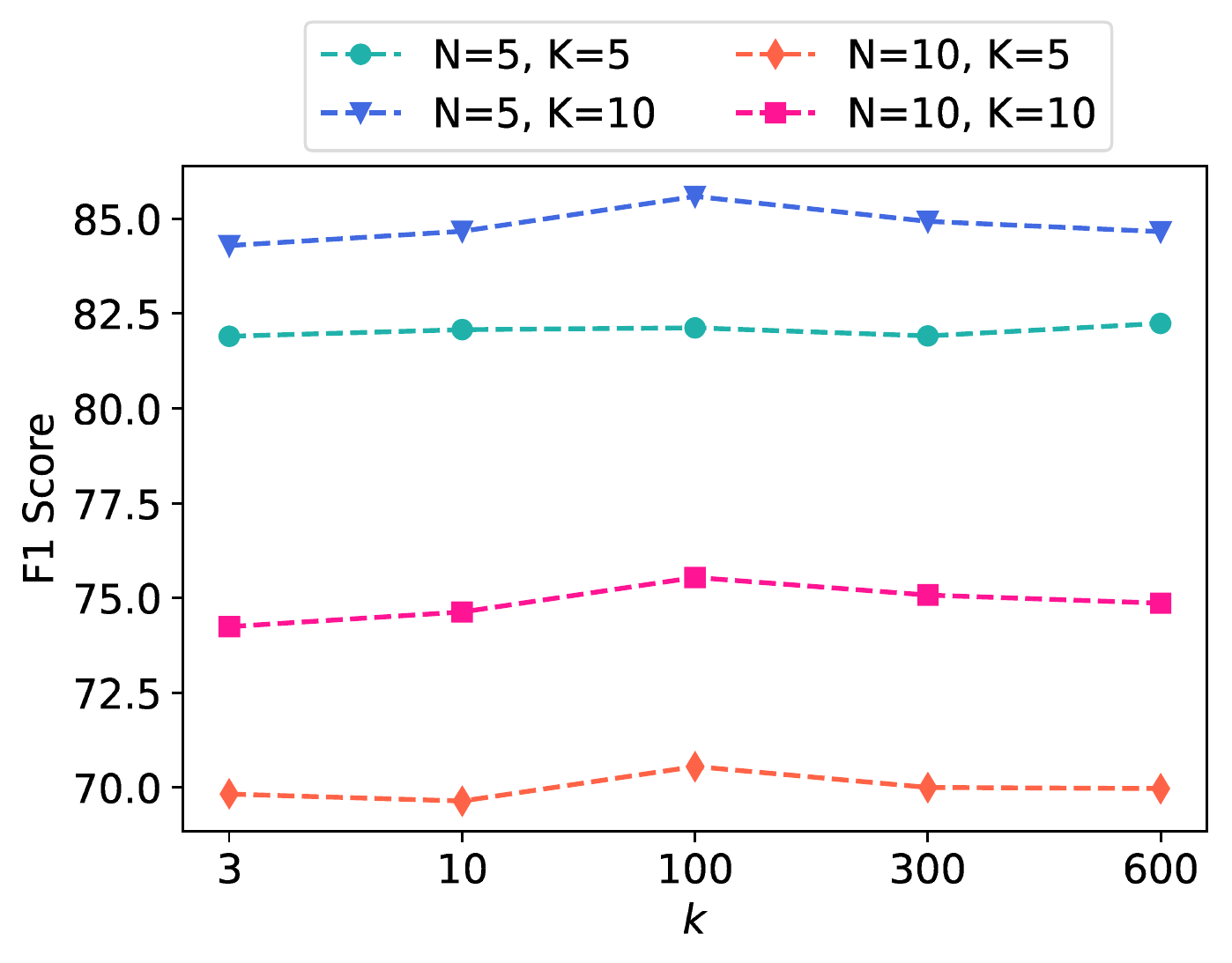}
			\label{ablation_auc}
		}
		\caption{The performance of LPN (w,w) with different $k$ values on the validation set of FewAsp.}
		\label{ablation_factor}
\end{figure}

\begin{figure*}[t]
	\centering
    \subfigure[Pic.1: LPN (o, o).]{
		\includegraphics[height=3.5cm,width=4.2cm]{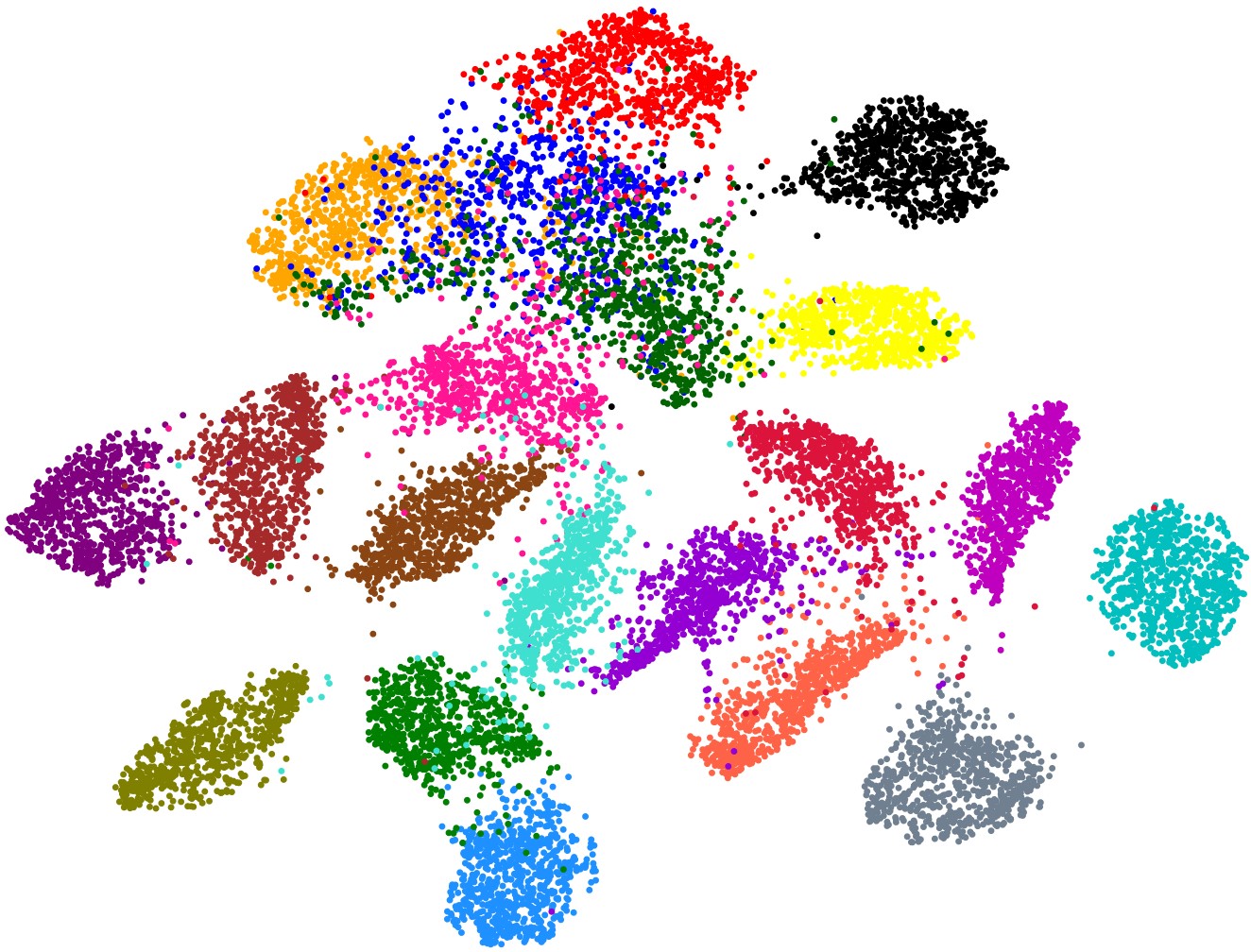}
	}\hspace{8mm}
	\subfigure[Pic.2: LPN (w, o).]{
		\includegraphics[height=3.5cm,width=4.2cm]{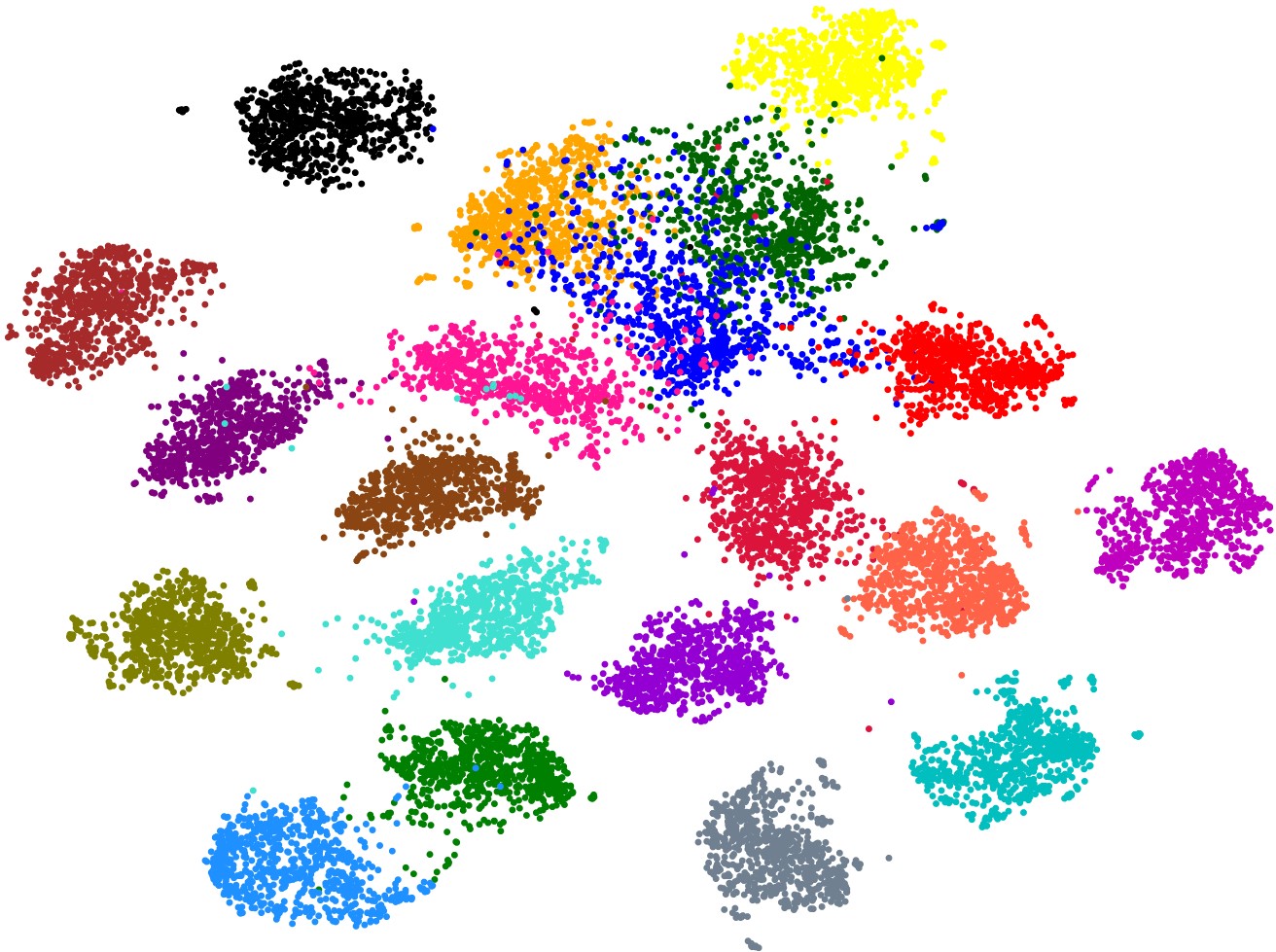}
	}\hspace{8mm}
	\subfigure[Pic.3: LPN (w, w).]{
		\includegraphics[height=3.5cm,width=4.2cm]{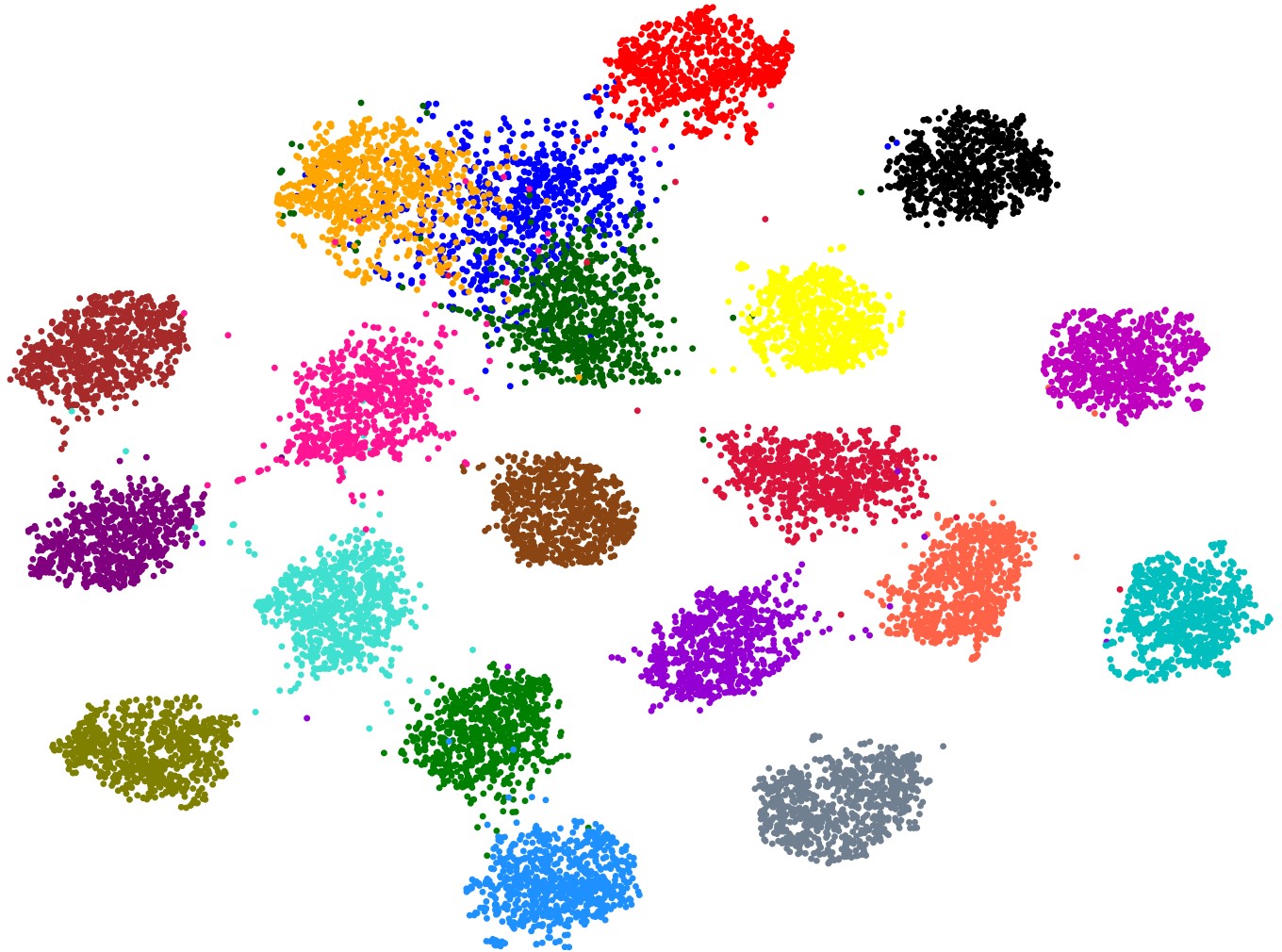}
	}
	\caption{Visualization of prototype embeddings obtained from LPN (o, o), LPN (w, o) and LPN (w, w) respectively.}
	\label{scl}
\end{figure*}

\subsection{Implementation Details}
\noindent\textbf{Evaluation Metric.} We follow \cite{DBLP:conf/acl/HuZGXGGCS20} to adopt two widely used metrics Area Under Curve (AUC) and macro-F1 score to evaluate the performance. 

\noindent\textbf{Parameter Settings.} For all experiments, we use the pre-trained language model Bert \cite{DBLP:conf/naacl/DevlinCLT19} to encode each word (token). Inspired by \cite{DBLP:journals/corr/abs-1911-03090}, we freeze the first 6 layers of Bert and fine-tune the final 6 layers. For the model parameters, we set $d=768$, $d'=256$, $R=4$ and $k=100$ consistently. For the loss function, we set $\tau = 0.1$, $\lambda = 0.1$ and $\gamma=0.01$ consistently and use AdamW \cite{DBLP:conf/iclr/LoshchilovH19} optimizer with the initial learning rate 1e-5. For these parameters, we use the grid searching strategy and validation set to determine them. Take parameter $k$ as an example. Figure \ref{ablation_factor} shows the performance of LPN (w,w) with different $k$ values on the validation set of FewAsp. It can be seen that when $k$ increases from 3 to 600, the performance improves at first and then drops, so we set $k=100$ in experiments. Table~\ref{parameter} summarizes the main hyperparameters of our model.

\subsection{Result Analysis}
We perform experiments with 5/10-way and 5/10-shot settings on FewAsp (single), FewAsp (multi) and FewAsp three datasets. All reported results are from 5 different runs, and in each run the results are averaged over 600 test episodes. Table \ref{tab.single}, \ref{tab.multi} and \ref{tab.all} show the experimental results for FewAsp (single), FewAsp (multi) and FewAsp respectively. The baseline
results are taken from \cite{DBLP:conf/acl/HuZGXGGCS20} and the best results are highlighted in bold. From the results, we could make the following observations.

(1) LPN performs much better than other baselines. Specifically, in terms of AUC, LPN improves upon the most competitive baseline Proto-AWATT by 1.53\%-2.13\%, 2.66\%-4.71\% and 1.87\%-3.30\% on FewAsp (single), FewAsp (multi) and FewAsp respectively. In terms of macro-F1 score, LPN improves upon Proto-AWATT by 5.86\%-9.12\%, 5.11\%-8.39\% and 4.74\%-6.85\% on FewAsp (single), FewAsp (multi) and FewAsp respectively. The reason is that LPN leverages label description as
auxiliary knowledge to learn more discriminative prototypes, integrates with contrastive learning to obtain better embeddings and uses a more effective multi-label inference module to accurately compute the aspect count.

(2) For all the methods, the results on FewAsp (multi) are a little worse than those on FewAsp (single) and FewAsp. The reason is that FewAsp (multi) consists of a large amount of sentences with multiple aspects, which increases the complexity of the dataset greatly. However, the proposed LPN can still achieve the best performance compared with other baselines, which further demonstrates the superiority of LPN in dealing with more complex multi-label tasks.

\subsection{Ablation Study}
\noindent \textbf{Label-enhanced Prototypes.}
To verify the effectiveness of label-enhanced prototypes, we make the ablation study. The results are shown in Table \ref{tab.single}, \ref{tab.multi} and \ref{tab.all}. LPN (o,o) means the LPN model without label enhancement and contrastive learning, and LPN (w,o) means the LPN model with label enhancement and without contrastive learning. It is easy to find that LPN (w,o) always performs much better than LPN (o,o) in all cases, which validates the effectiveness of the label-enhanced prototypes. The reason is that label text descriptions contain lots of aspect-relevant semantic information, which is highly conducive to obtain more discriminative prototypes.
	
\noindent \textbf{Contrastive Learning.}
We also make the ablation study for the module of contrastive learning. The results are shown in Table \ref{tab.single}, \ref{tab.multi} and \ref{tab.all}. LPN (w,o) means the LPN model with label enhancement and without contrastive learning, and LPN (w,w) means the LPN model with label enhancement and contrastive learning. We can observe that in most cases LPN (w,w) performs better than LPN (w,o). This is because that constrastive learning module can push samples in the same class close and samples in different classes further apart, thus obtaining better sentence embeddings.

\subsection{Visualization}
To better observe how the embeddings change with label-enhanced prototypes and contrastive learning, we sample 3000 episodes from test set of FewAsp (multi) in 5-way-5-shot setting, and then use t-SNE \cite{van2008visualizing} to visualize the prototype embeddings obtained from LPN (o,o), LPN (w,o) and LPN (w,w). Note that we originally intend to visualize the sentence embeddings, but each sentence may contain multiple aspects which is difficult to distinguish by color. As each prototype is associated with unique aspect and is generated by the corresponding intra-class sentences, it can represent the sentence embedding to some extent. Figure \ref{scl} gives the visualization result of prototype embeddings obtained from LPN (o,o), LPN (w,o) and LPN (w,w). Prototypes (data points) with the same color contains the same aspect. It is easy to find that the distribution generated by LPN (o,o) has a lot of overlaps. The label enhancement in LPN (w,o) can help to separate the embeddings to some extent. The contrastive learning can further guarantee that the embeddings from same class are pulled together and the embeddings from different classes are pushed apart.

\section{Conclusion}
In this paper, we propose a label-enhanced prototypical network (LPN) to deal with multi-label few-shot aspect category detection. To learn more discriminative prototypes, LPN adopts the label text description as auxiliary knowledge to retain aspect-relevant information while eliminating the negative effect triggered by irrelevant aspects. To obtain better sentence embeddings to facilitate the aspect category detection task, LPN introduces contrastive learning to reduce intra-class discrepancy and enlarge the inter-class difference among sentence embeddings. Extensive experiments on three real-world datasets show that LPN outperforms the state-of-the-art methods by a large margin. In future work, we plan to investigate the theoretical underpinnings of our approach and extend our model to other multi-label few-shot scenarios like intent detection in dialogue systems.

\begin{acks}
The authors are grateful to the anonymous reviewers for their valuable comments and suggestions. This work was supported by National Natural Science Foundation of China (No. 62106035, 61876028), and Fundamental Research Funds for the Central Universities (No. DUT20RC(3)040, DUT20RC(3)066).
\end{acks}

\bibliography{sample-base}
\bibliographystyle{ACM-Reference-Format}

\end{document}